\documentclass[10pt,conference]{ieeeconf} 

\IEEEoverridecommandlockouts

\let\labelindent\relax
\usepackage{enumitem}
\usepackage{balance}
\usepackage[%
    style=ieee,
    sorting=none,
    natbib=true,  
    backend=biber,
    sortcites=true,
    doi=false,
    url=false,
    giveninits=true,
    maxcitenames=4, 
    minbibnames=3, 
    maxbibnames=4, 
    hyperref
]{biblatex}
\bibliography{ref/external,ref/swarm.bib}

\usepackage{xurl}
\usepackage{setspace}
\usepackage[skip=0pt]{caption}

\usepackage{graphicx} 
\usepackage{float}
\usepackage{bm}
\usepackage{mathtools}
\usepackage{mwe}
\usepackage{subcaption}

\usepackage{amsmath, amsthm, amssymb}

\usepackage[usenames,dvipsnames]{xcolor}
\usepackage{accents}
\usepackage{hyperref}
\usepackage[capitalize]{cleveref}
\usepackage{siunitx}
\usepackage[acronyms,shortcuts]{glossaries}

\title{\LARGE \bf Robust Forecasting for Robotic Control: A Game-Theoretic Approach}

\author{Shubhankar Agarwal$^{1}$, David Fridovich-Keil$^{2}$, and Sandeep P. Chinchali$^{1}$
\thanks{$^{1}$Departments of Electrical  and Computer Engineering and $^{2}$ Aerospace Engineering \& Engineering Mechanics, The University of Texas at Austin, {\tt\small \{somi.agarwal, dfk, sandeepc\}@utexas.edu}}}

\crefname{problem}{Problem}{Problems}
\crefname{assumption}{Assumption}{Assumptions}
\crefname{remark}{Remark}{Remarks}
\crefname{proposition}{Proposition}{Propositions}
\crefformat{equation}{(#2#1#3)} 

\newtheorem{remark}{Remark}
\newtheorem{definition}{Definition}
\newtheorem{proposition}{Proposition}
\newtheorem*{proposition-non}{Proposition}
\newtheorem{problem}{Problem}

\newcommand{\forecaster}{f}
\newcommand{\adversary}{a}
\newcommand{\dynamics}{g}

\newcommand{\paramforcaster}{\theta_{f}}
\newcommand{\paramcontroller}{\theta_{c}}
\newcommand{\paramadversary}{\theta_{a}}
\newcommand{\paramspaceforcaster}{\Theta_f}
\newcommand{\paramspaceadversary}{\Theta_a}

\newcommand{\historyfull}{s_{H}}
\newcommand{\futurepredfull}{\hat{s}_{F}}
\newcommand{\futuregtfull}{s_{F}}
\newcommand{\history}{\mathbf{s_{H}}}
\newcommand{\futurepred}{\mathbf{\hat{s}_{F}}}
\newcommand{\futuregt}{\mathbf{s_{F}}}
\newcommand{\historyadv}{\mathbf{s}^{\text{adv}}_{\mathbf{H}}}

\newcommand{\controlcost}{J^C}

\newcommand{\traininput}{x}
\newcommand{\trainoutputgt}{y}
\newcommand{\trainoutputpred}{\hat{y}}
\newcommand{\dataset}{\mathcal{D}}
\newcommand{\datasetall}{\dataset = \{\left(x, y\right)\}_{i=1}^{N}}

\newcommand{\datasetorigtrain}{\dataset_{\mathrm{orig}}^{\mathrm{train}}}

\newcommand{\datasetaddtrain}{ \mathcal{D}_{\mathrm{add}}^{\mathrm{train}} }

\newcommand{\datasetoodtest}{\dataset_{\mathrm{ood}}^{\mathrm{test}}}

\newcommand{\datasetrandtrain}{\dataset_{\mathrm{rand}}^{\mathrm{train}}}

\newcommand{\gamecostsymb}{\mathcal{J}}
\newcommand{\finalgamecost}[2]{\gamecostsymb(#1, #2)}

\newcommand{\Original}{\textsc{Original}}
\newcommand{\DataAdded}{\textsc{Data Added}}
\newcommand{\Random}{\textsc{Random}}
\newcommand{\Robust}{\textsc{Robust}}

\glsdisablehyper
 
\newacronym{ood}{OoD}{Out-of-Distribution}
\newacronym[longplural=Long-Short Term Memories,plural=LSTMs]{lstm}{LSTM}{Long-Short Term Memory}
\newacronym{gne}{GNE}{global Nash equilibrium}
\newacronym{lne}{LNE}{local Nash equilibrium}

\begin{document}

\maketitle

\begin{abstract}
Modern robots require accurate forecasts to make optimal decisions in the real world. For example, self-driving cars need an accurate forecast of other agents' future actions to plan safe trajectories. Current methods rely heavily on historical time series to accurately predict the future. However, relying entirely on the observed history is problematic since it could be corrupted by noise, have outliers, or not completely represent all possible outcomes. 
To solve this problem, we propose a novel framework for generating robust forecasts for robotic control. In order to model real-world factors affecting future forecasts, we introduce the notion of an adversary, which perturbs observed historical time series to increase a robot's ultimate control cost. 
Specifically, we model this interaction as a zero-sum two-player game between a robot's forecaster and this hypothetical adversary. We show that our proposed game may be solved to a local Nash equilibrium using gradient-based optimization techniques. Furthermore, we show that a forecaster trained with our method performs $30.14\%$ better on out-of-distribution real-world lane change data than baselines.

\end{abstract}
\section{Introduction}

Robots deployed in the real world rely on accurate forecasts of the future to make reliable decisions amidst uncertainty. For example, an autonomous vehicle must forecast the trajectory of cars in an adjacent lane in order to decide when and how to change lanes. The problem of accurate forecasting is likewise essential in other large-scale and safety-critical systems, such as electric grids and communications networks, and transportation systems.

\begin{figure}[tp]
    \centering
    \includegraphics[width=\columnwidth]{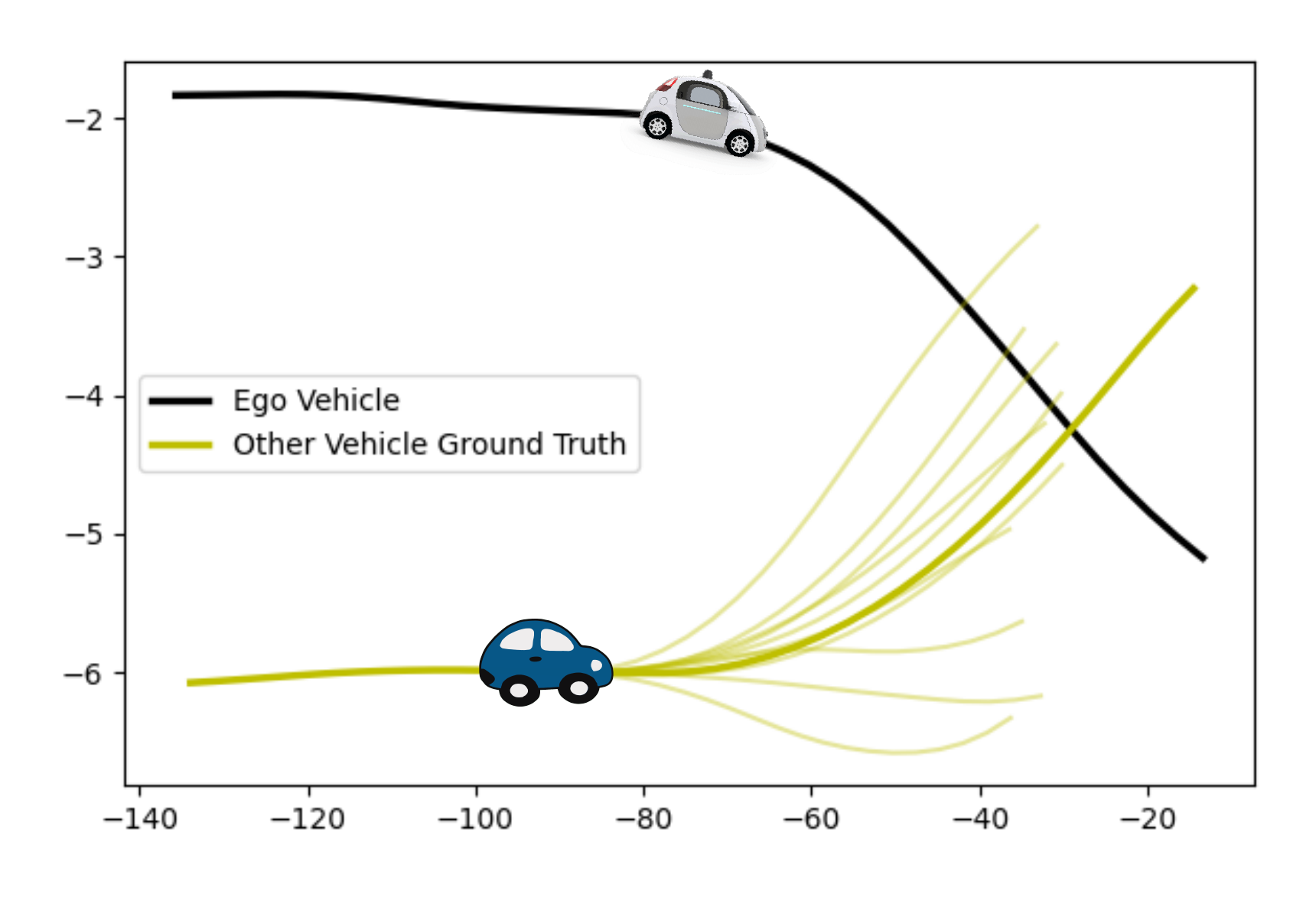}
    \includegraphics[width=\columnwidth]{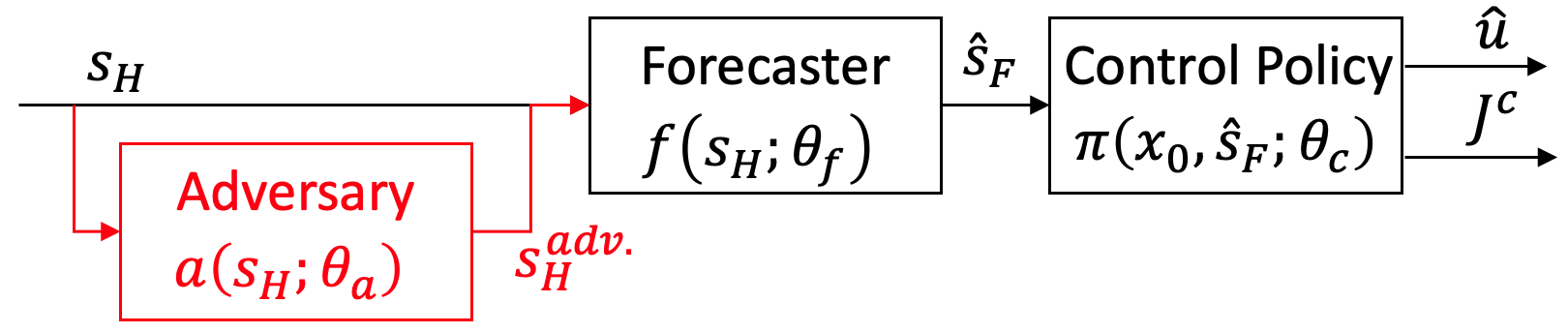}
\vspace{-0.5em}
\caption{\small{\textbf{Motivating Example (top):} Lane change scenario involving an autonomous ``ego'' car (white) and another, potentially human-operated car (blue). The ego vehicle needs to plan its future trajectory given historical observations of the other vehicle (bold yellow line). Other highlighted yellow lines show several possible trajectories for the blue car, some resulting in faster or even unsuccessful lane changes. Even though the ego vehicle forecaster observes only a single scenario (bold yellow line), it needs to be robust to these other possible outcomes. \textbf{Architecture (bottom):} Our robotic system consists of a forecaster, hypothetical adversary, and controller.}} 
\label{fig:motivating_example}
\vspace{-1.5em}
\end{figure}

However, forecasting the future is one of the most challenging problems in machine learning. Current methods such as \cite{lim2021deep, Ivanovic2021traj, Makridakis1976survey} rely heavily on historical time series to accurately predict the future. However, completely relying on the observed history is problematic since it could be corrupted by sensor noise, have outliers, or not completely represent all possible outcomes. For example, in the case of self-driving cars, the observed history could be corrupted by noise due to sensor imprecision or outliers due to human labeling errors. Moreover, the observed historical training dataset might be incomplete---in the case of lane-change maneuvers, we might only observe that if one car slows down, the other car completes the lane change first (see \cref{fig:motivating_example}). However, we might not observe outlier behavior where a car slows down and subsequently speeds up. Systems deployed in the real world need to be robust to these problems and generate forecasts which consider these situations. 
Standard data engineering practices, such as collecting more targeted data or adding random noise to existing data, are either expensive or not reflective of important outliers, respectively. 

We propose a novel framework for generating robust forecasts for optimal decision-making in robotics. We consider a system, consisting of a forecaster and a controller, represented in \cref{fig:motivating_example}. The forecaster observes historical time series data and makes a prediction which the controller uses to determine optimal future actions. For example, a self-driving car's forecaster predicts the future behavior of a human-driven car and the controller subsequently determines the self-driving car's optimal lane change maneuver in response. To model real-world factors affecting the future forecasts, we introduce a notion of an adversary, which perturbs the historical time series, leading to inaccurate forecasts. As such, the forecaster and adversary play a game where the adversary tries to maximize its reward by perturbing the forecaster's input, while the forecaster minimizes its cost by performing well on the adversarially perturbed inputs. 
Motivated by this observation, our contributions are as follows.




\textbf{Contributions:} First, we formulate the problem of robust forecasting for optimal control as a two-player, zero-sum game. Recent advances in numerical game theory provide gradient-based algorithms which are guaranteed to find \acp{lne} in such problems. Second, we show the benefits of a robustly-trained forecaster on two different datasets, including a $30.14\%$ improvement in performance on out-of-distribution real-world lane change data as compared to baseline forecasters.

\section{Related Work}
\textbf{Forecasting:}
Forecasting ego-vehicle trajectories is primarily studied via data-driven and probabilistic techniques. Data-driven trajectory forecasting approaches, such as \cite{bartoli2017context, alahi2016social, fernando2018soft, lee2017desire, giuliari2020transformer, singh2023trajectory, singh2023trajectory}, treat both single and multi-trajectory forecasting problems within a framework of temporal regression, using models such as \acp{lstm} \cite{Hochreiter1997Long} and Transformers \cite{Vaswani2017Attention}. 
These methods typically do not account for outliers in the dataset or \ac{ood} scenarios. In contrast, probabilistic forecasting models such as \cite{Rhinehart_2019_ICCV, Varadarajan2022Multipath, Song2022Learning, Salzmann2020Trajectron, Schmerling2018multimodal} output a distribution of possible future trajectories, and can thus account for outliers and multimodality. Still, these techniques have difficulty generalizing to \ac{ood} scenarios since the distributions they learn are based only upon fixed training data. 
In particular, none of these approaches consider robustness to adversarial distribution shifts. 

\textbf{Adversarial Machine Learning for Control:} 
Adversarial modeling is widely studied in the context of robust control and decision making \cite{baser1998dynamic,Esfahani2015Data, sinha2018certifiable, volpi2018generalizing}. 
For example, the introduction of an adversary to improve a machine learning model's robustness and generalization capabilities is commonly studied in vision and classification tasks such as \cite{madry2018towards, pmlr-v80-wong18a, NEURIPS2019_e2c420d9}.
However, existing works in robust machine learning do not consider the downstream implications on control performance for robotic decision problems.
Works which do study adversarial attacks in control systems consider settings in which control decisions are directly perturbed by adversarial inputs \cite{GenAdvControl, ICMLControlAdv}.
They do not consider a more general setting in which inputs are provided to a forecaster which then invokes an internal model-based controller, as we consider in this work.
Additionally, \cite{li2022adversarial} design adversarial attacks which increase a cost metric as well as violate state and control constraints. However, they did not study how to exploit these adversarial attacks to design an improved, robust decision process for robotic applications. Recent work, \cite{agarwal2022synthesizing} employs a similar problem  as our method but does not consider the game formulation and thus is not able to provide theoretical insights on convergence or robustification abilities.

\textbf{Game Theory:}
Learning in noncooperative settings such as games exposes significant challenges regarding convergence, stability of desired solutions, etc.
For example, these problems are well known in the context of generative adversarial networks \cite{Goodfellow2014Generative}, which are notoriously difficult to train \cite{arjovsky2017towards}.
However, recent advances in numerical game theory \cite{Mazumdar2019OnFL, Mazumdar2020On, Raghunathan2019Game,fiez2019convergence,fiez2021global,zheng2022stackelberg,laine2021computation} provide important steps forward on several of these fronts.
In particular, these algorithmic advances ensure that local equilibrium solutions to the games considered in this work can be found reliably.
\section{Problem Formulation}

We now describe the interplay between the robot's forecaster, controller, and the adversary (see \cref{fig:motivating_example}). They all operate in discrete time steps $t$ for a horizon of $T$ steps. 

\textbf{Forecaster:} The forecaster observes a time series $s_T \in \mathbb{R}^{p \times T}$, drawn from a data distribution $\mathcal{D}$, denoted by $s_{T} \sim \mathcal{D}$. The forecaster, $\forecaster : \mathbb{R}^{p \times H} \rightarrow \mathbb{R}^{p \times F} $, maps the time series of past $H$ measurements, denoted by $ \historyfull$, to a time series of future $F$ measurements, denoted by $\futurepredfull$. The hat notation, $\hat{s}$, denotes predicted values of the forecaster, and $\futuregtfull$ denotes the ground-truth time series. The forecaster is a learned module parameterized by $\paramforcaster \in \paramspaceforcaster$, where $\paramspaceforcaster$ is the space of all possible parameters. To simplify notation, we use bold variables to define the full time series, i.e., the time series of past $H$ measurements as $\history$.

\textbf{Control Policy:} The control policy $\pi : \mathbb{R}^n \times \mathbb{R}^{p \times F} \rightarrow \mathbb{R}^m$ maps the state of system, $x_t \in \mathbb{R}^n$, and time series of future $F$ measurements, $\futuregt$, to an optimal control $u_t \in \mathbb{R}^m$. We denote the state and control constraint sets by $\mathcal{X}$ and $\mathcal{U}$ respectively. The robot dynamics, $\dynamics$, are given by: $x_{t+1} = \dynamics(x_t, u_t), \forall t \in \left\{ 0, \ldots, T - 1 \right\}$. Ideally, control policy $\pi$ chooses a decision $u_t$ at time $t$ based on fully-observed state $x_t$ and perfect knowledge of exogenous input $\futuregt$: $u_t = \pi(x_t, \futuregt; \paramcontroller)$, where $\paramcontroller$ are control policy parameters. 
For example, if the control policy $\pi$ were derived from solving a Linear Quadratic Regulator (LQR) problem, $\paramcontroller$ represents a linear state feedback matrix derived from the unique solution to the discrete algebraic Riccati equation.
However, in practice, given a possibly perturbed forecast $\futurepred$, it will enact a control denoted by $\hat{u}_t = \pi(x_t, \futurepred; \paramcontroller)$, which depends on the forecaster parameters $\paramforcaster$ via the forecast $\futurepred$.  



\textbf{Control Cost:} The main objective is to minimize the control cost $\controlcost$, which depends on initial state $x_0$ and controls $\hat{u}_{0:T-1}$. The control cost $\controlcost$ is a sum of stage costs $c(x_t, s_t, \hat{u}_t)$ and terminal cost $c_T(x_T, s_T)$, i.e., $\controlcost(\mathbf{\hat{u}}; x_0, \mathbf{s}) = \sum_{t=0}^{T -1} c(x_t, s_t, \hat{u}_t) + c_T(x_{T}, s_{T})$. 

\textbf{Adversary:} In order to account for the measurement noise in the inputs of the forecaster and improve generalization to out-of-distribution data, we introduce an adversary in our robotic system which can perturb the inputs of the forecaster. The adversary can be viewed as a hypothetical, virtual agent that perturbs historical training data in order to corrupt forecasts and ultimately make control performance worse. More concretely, the adversary is defined as the map $\adversary: \mathbb{R}^{p \times H} \rightarrow \mathbb{R}^{p \times H}$, which takes as input the time series of past $H$ measurements, $\history$, and outputs an adversarially perturbed version of the past $H$ measurements denoted as $\historyadv$. The adversary is parameterized by $\paramadversary \in \paramspaceadversary$, where $\paramspaceadversary$ is the space of all possible parameters. Additionally, in order to restrict the power of the adversary, we penalize the adversary quadratically for large perturbations: $\lVert \history - \historyadv \rVert_2^2$.

\textbf{Overall Cost:} 
The system operates as follows. First, the adversary (with parameters $\paramadversary$) perturbs the historical time series, $\historyadv$, given the actual history of time series $\history$. Then, the forecaster observes the adversary's perturbed history, $\historyadv$, and predicts the future state of the system, $\futurepred$ \cref{eq:forward}. Finally, given the predicted forecast  $\futurepred$ and the ground-truth forecast $\futuregt$, we calculate the corresponding optimal controls $\hat{\mathbf{u}}$ and $\mathbf{u}$, respectively, for the system using the control policy \cref{eq:control}: 
\begin{subequations} \label{eq:system}
\begin{align}
    \historyadv = \adversary(\history; \paramadversary),& \hspace{1em} \futurepred = \forecaster(\historyadv; \paramforcaster). \label{eq:forward} \\
    \hat{\mathbf{u}} = \pi \left( x_0, \futurepred; \paramcontroller \right),& \hspace{1em} \mathbf{u} = \pi \left( x_0, \futuregt; \paramcontroller \right). \label{eq:control}
\end{align}
\end{subequations}
Thus equipped, we calculate the overall cost \cref{eq:final_cost}. The first term calculates the additional cost incurred by using predicted forecasts $\futurepred$ instead of true forecast $\futuregt$. This term models the change in states $x$ and controls $u$ given the errors in the prediction of the future time series. The second term $\lVert \futuregt - \futurepred \rVert_2^2$ penalizes deviations of the forecaster's future time series predictions and the ground-truth forecast. The third term $\lVert \history - \historyadv \rVert_2^2$ controls the adversary's power by penalizing it for making large perturbations from the historical time series. The hyper-parameters $\lambda_f$ and $\lambda_a$ control the relative importance of the respective costs:
\begin{equation} \label{eq:final_cost}
\begin{aligned}
     J\left(\cdot \right) &= \controlcost\left(\hat{\mathbf{u}} ; x_{0}, \futuregt \right)-\controlcost\left(\mathbf{u} ; x_{0}, \futuregt \right) \\
    &\hspace{0.5cm} + \lambda_f \lVert \futuregt - \futurepred \rVert_2^2  - \lambda_a \lVert \history - \historyadv \rVert_2^2.
\end{aligned}
\end{equation}


For clarity, we introduce the following compact notation for this cost---$\finalgamecost{\paramforcaster}{\paramadversary}$ is the final overall cost $J\left(\mathbf{u}, \hat{\mathbf{u}}, \futuregt, \futurepred ; x_{0} \right)$ with fixed parameters $\paramforcaster$ and $\paramadversary$. The total cost depends on the controller and forecaster parameters via controls $\mathbf{u}$ and $\hat{\mathbf{u}}$ and the forecast $\futuregt$. Importantly, the controller's parameters $\paramcontroller$ are determined \emph{implicitly} as the solution to the aforementioned optimal control problem, for each choice of $\paramadversary$ and $\paramforcaster$. 
Therefore, we may focus our attention on optimizing only the forecaster and the adversary parameters. Having defined the information flow in our robotic system, we now formalize the problem addressed in this paper.
\begin{problem}[Adversarially-Robust Control] \label{problem:robust_against_adversary} 
Given a forecaster $\forecaster$ and an adversary $\adversary$, solve the min-max problem:
\vspace{-1em}
\begin{subequations} \label{eq:prob_robust_against_adversary}
\begin{align}
\min_{\paramforcaster} \max_{\paramadversary} & \hspace{1em} \finalgamecost{\paramforcaster}{\paramadversary} \label{eq:prob_robost_cost} \\ 
\text{subject to:~ } \historyadv &= \adversary(\history; \paramadversary),\, \futurepred = \forecaster(\historyadv; \paramforcaster), \label{eq:prob_forward} \\
                   \hat{\mathbf{u}} &= \pi \left( x_0, \futurepred; \paramcontroller \right),\, \mathbf{u} = \pi \left( x_0, \futuregt; \paramcontroller \right). \label{eq:prob_control}
\end{align}
\end{subequations}
Specifically, we aim to find a saddle point of \cref{eq:prob_robust_against_adversary} in which the order of the minimum and maximum does not matter. 
\end{problem}

In \cref{problem:robust_against_adversary}, the adversary is trying to maximize the overall control cost \cref{eq:prob_robost_cost} by perturbing the original past $H$ measurements, $\history$ \cref{eq:prob_forward}. In contrast, the forecaster is trying to minimize the overall cost, \cref{eq:prob_robost_cost}, by forecasting the future $F$ measurements, $\futurepred$ \cref{eq:prob_forward}. Intuitively, this problem captures how to find a forecaster that is robust to adversarial perturbations for reliable robotic decision-making.

\section{Approach}

We observe that \cref{problem:robust_against_adversary} is a two-player, zero-sum game, and seek both forecaster and adversary parameters which are in equilibrium. 


\subsection{Characterizing the Robust Forecasting Game}

Here we provide a precise characterization of this game, and introduce key solution concepts.

\textbf{Player 1 (Forecaster):} The forecaster's goal is to predict relevant future system states, despite worst-case perturbations of the history by the adversary.
In the lane changing example of \cref{fig:motivating_example}, for example, this is the future trajectory of the ego car. Thus equipped, the forecaster seeks parameters $\paramforcaster^*$ which minimize the overall cost in  \cref{problem:robust_against_adversary} despite worst-case adversarial parameter selection, $\paramadversary^*$.



\textbf{Player 2 (Adversary):} The adversary's goal is to provide a perturbed history to the forecaster such that the predicted future time series by the forecaster incurs a higher overall cost. In particular, for any fixed choice of forecaster parameter $\paramforcaster$, it seeks corresponding parameters $\paramadversary$ which maximize the overall cost in \cref{problem:robust_against_adversary}, $\gamecostsymb(\cdot, \paramforcaster)$.

We are now ready to describe relevant solution concepts in \cref{problem:robust_against_adversary}.

\begin{definition}[\Acl{gne}] \label{def:global_nash_equilibrium}
\cite[Defn. 2.1]{baser1998dynamic}
A pair of actions, $\paramforcaster^*$ and $\paramadversary^*$, are a \ac{gne} of a game if for all $\paramforcaster$ and $\paramadversary$ in $\paramspaceforcaster \times \paramspaceadversary$: 
    $$\finalgamecost{\paramforcaster^*}{\paramadversary} \leq \finalgamecost{\paramforcaster^*}{\paramadversary^*} \leq \finalgamecost{\paramforcaster}{\paramadversary^*}\,.$$
\end{definition}

\begin{definition}[\Acl{lne}] \label{def:local_nash_equilibrium} 
\cite[Defn. 1]{ratfliff2016On} Let $\|\cdot\|$ denote a vector norm.
    A pair of actions, $\paramforcaster^*$ and $\paramadversary^*$, are a \acf{lne} of cost function $\gamecostsymb$ if there exists an $\epsilon > 0$ such that for any parameters $\paramforcaster$ and $\paramadversary$ satisfying $\lVert \paramforcaster - \paramforcaster^* \rVert \leq \epsilon$ and $\lVert \paramadversary - \paramadversary^* \rVert \leq \epsilon$, we have:
    $$\finalgamecost{\paramforcaster^*}{\paramadversary} \leq \finalgamecost{\paramforcaster^*}{\paramadversary^*} \leq \finalgamecost{\paramforcaster}{\paramadversary^*}\,.$$
\end{definition}

A \ac{gne} is a point in the space of game strategies where both players cannot change their respective strategy without achieving a less favorable outcome. A \ac{lne} is a point in the space of strategies where this property need only hold within a small neighborhood. \Cref{def:global_nash_equilibrium,def:local_nash_equilibrium} are standard solution concepts in the theory of smooth static games.

\Acp{lne} are characterized by the following first- and second-order optimality conditions.

\begin{proposition}[First-order Necessary Condition] \label{prop:first_order_necessary_condition}
    Assuming $\gamecostsymb$ is differentiable, any local Nash equilibrium satisfies $\nabla_{\paramforcaster}\finalgamecost{\paramforcaster}{\paramadversary}=0$ and $\nabla_{\paramadversary}\finalgamecost{\paramforcaster}{\paramadversary}=0$.
\end{proposition}

\begin{proposition}[Second-order Sufficient Condition] \label{prop:second_order_necessary_condition}
    Assuming $\gamecostsymb$ is twice-differentiable, any local Nash equilibrium satisfies $\nabla_{\paramforcaster\paramforcaster}^2 \finalgamecost{\paramforcaster}{\paramadversary} \succeq 0$, and $\nabla_{\paramadversary\paramadversary}^2 \finalgamecost{\paramforcaster}{\paramadversary} \preceq 0$.
\end{proposition}


\subsection{Training the Models} \label{sec:nn_models}
In this work, we use feedforward neural networks (NNs) to represent the forecaster and adversary models. Due to the nonconvexities in overall cost $\gamecostsymb$, we can at best guarantee that our proposed game will reach a \ac{lne}. More precisely, \cite{Mazumdar2020On} demonstrates that stochastic gradient descent methods do not necessarily converge to \ac{lne} in zero-sum games, but \cite{Mazumdar2019OnFL} proposes a new second-order gradient update rule that does guarantee convergence to a \ac{lne} if one exists. 
However, due to complexities and speed limitations of the second-order gradient update method, we follow standard practices in high-dimensional optimization and resort to an adaptive first-order method such as ADAM \cite{Kingma2015AdamAM}. 
Since we do not use the theoretically-motivated second-order technique of \cite{Mazumdar2019OnFL}, we take care to check the first- and second-order conditions of \cref{prop:first_order_necessary_condition,prop:second_order_necessary_condition} to ensure that we have found a \ac{lne}, as described later in Section  \ref{sec:data_schemes}.



\begin{remark}[Robustness to Adversarial Perturbations] \label{remark:robust}
 It is readily apparent that $\gamecostsymb(\paramadversary^*, \paramforcaster^*) \le \gamecostsymb(\paramadversary, \paramforcaster^*)$ when the cost function $\gamecostsymb(\cdot, \paramforcaster^*)$ is concave in the first argument.
 However, the examples considered in this work do not exhibit such concavity.
 Nevertheless, our experimental results demonstrate that, for \ac{lne} forecaster parameters $\paramforcaster^*$, the cost $\gamecostsymb(\cdot, \paramforcaster^*)$ is substantially robust to adversarial perturbations.
\end{remark}

\section{Experiments}

We now evaluate our method on two different scenarios. The first is a synthetic Autoregressive Integrated Moving Average (ARIMA) process \cite{Harvey1990ARIMA} generated by random parameters. In the second task, we use lane-change data from an autonomous driving scenario with human participants \cite{Schmerling2018multimodal}. The experiments aim to demonstrate that 1) our proposed game converges to a \ac{lne} and 2) the forecaster trained using our proposed method will be robust to \ac{ood} data.
We now discuss commonalities between both experiments.


\textit{Models:} Both the forecaster and adversary are NN models with two fully connected layers and ReLU activations.

\textit{Differentiable Model Predictive Control (MPC):} In both tasks, the control policy $\pi$ is the solution map of an MPC problem with quadratic costs and linear constraints. The forecaster provides the MPC controller with a future time series forecast $\futurepred$ to track and the current state, $x_0$. We use linear dynamics, $\dynamics$, in our MPC formulation. Specifically, we used linear 1D dynamics for the ARIMA experiment and second-order linear dynamics for the lane-change experiment. For both the experiments, the stage cost $c(x_t, s_t, \hat{u}_t)$ and terminal cost $c_T(x_{T}, s_{T})$ in the control cost $J^C$ are quadratic in the state $x_T$ and controls $\hat{u}_T$. Specifically, the terminal cost is $c_T(x_T, s_T) = \left(x_T - s_T \right)^{\top} Q \left(x_T - s_T \right)$ and the stage cost is $c(x_t, s_t, \hat{u}_t) = \left(x_t - s_t \right)^{\top} Q \left(x_t - s_t \right) + {\hat{u}_t}^{\top} R {\hat{u}_t}$, where $Q$ and $R$ are positive definite matrices.  The robot actuator constraints which are described by intervals along each axis, i.e. $u_\text{min} \leq u_t \leq u_{\text{max}}$. Likewise, we presume that states are also constrained to lie within an axis-aligned box: $ x_{\text{min}} \leq x_t \leq x_{\text{max}}$. While training the forecaster and the adversary, we require gradients of the control policy with respect to the forecaster and the adversary parameters. 
To do so, we use the \textsc{cvxpylayers} Pytorch library \cite{cvxpylayers2019}, which allows us to backpropagate derivative information through convex optimization problems and thereby train both the forecaster and adversary end-to-end.

\begin{figure*}[ht]
    \centering
    \includegraphics[width=1.0\columnwidth]{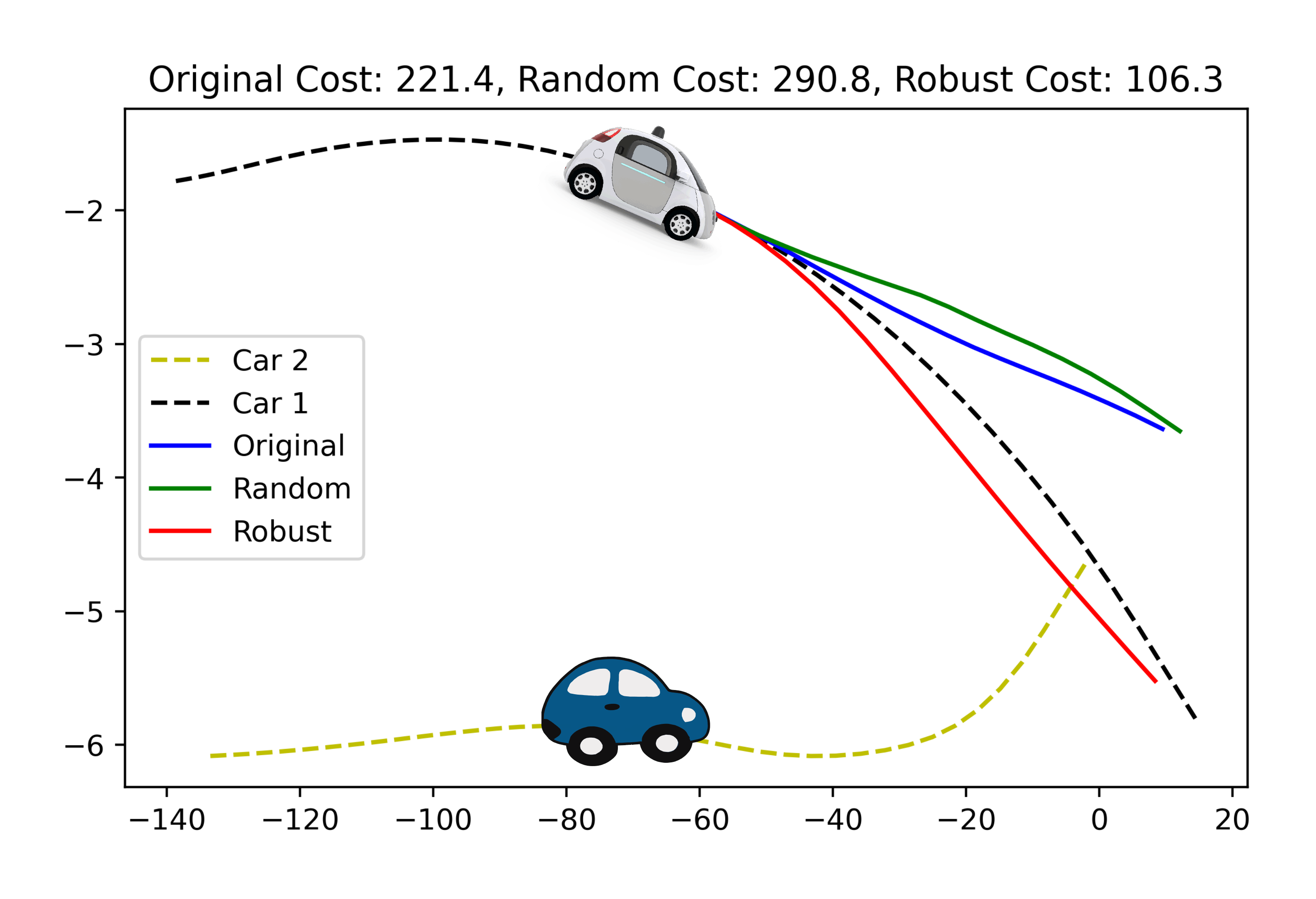}
    \includegraphics[width=1.0\columnwidth]{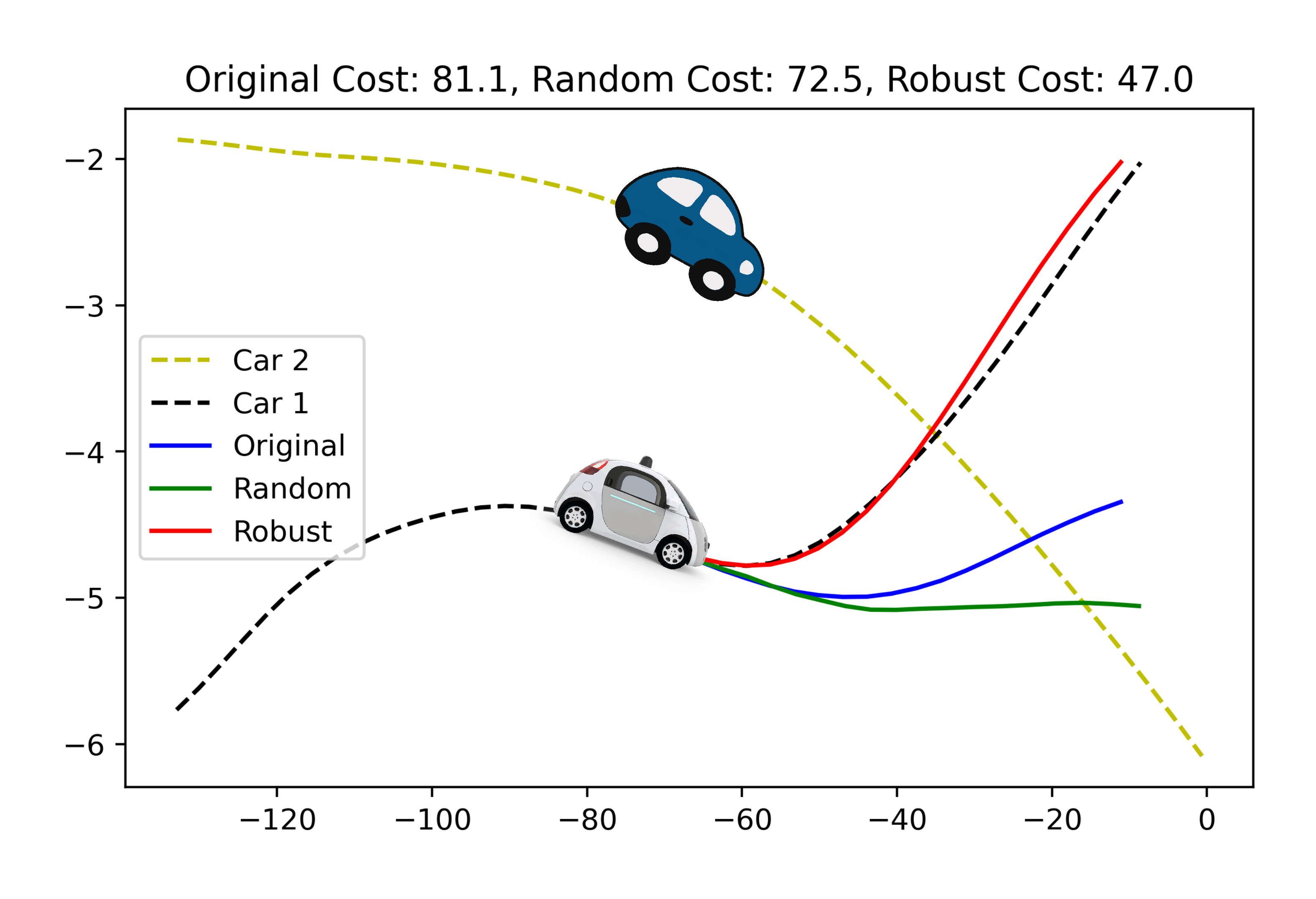}
    \caption{\small{\textbf{Examples of our \Robust \space scheme's performance on OoD scenarios:}
    We show two scenarios from the OoD test dataset. The axes are the scaled Cartesian coordinates of the vehicles. We compare the forecasted lane-change trajectories of \Original \space (blue), \Random \space (green), and \Robust \space (red) training schemes. The black and yellow dashed lines show the ground-truth trajectories of the ego-vehicle and the other vehicle respectively. All the forecasters were given the historical trajectories of both the vehicles before their current locations. We additionally show the control cost of all forecasted trajectories on the top. The key takeaway is that our \Robust \space scheme is able to generate trajectories closer to the ground-truth in \ac{ood} scenarios and thus is able to generalize better to unseen scenarios.}}
    \label{fig:adv_scenarios}
\vspace{-1.2em}
\end{figure*}

    

\subsection{Datasets and Benchmark Algorithms} \label{sec:data_schemes}
In the experiments, all time series forecasts are a tensor instead of a vector. For example, the historical time series is $\history \in \mathbb{R}^{N \times p \times H}$, where $N$ is the number of individual time series, $p$ is the dimension and $H$ is the horizon of the time series. We collect several examples of these time series tensors in a dataset, which we use to train the forecaster and the adversary. A dataset contains $N$ tuples of inputs $\traininput$ and labels $\trainoutputgt$ denoted by $\datasetall$. In each tuple, $\traininput = \left\{ \history, x_0 \right\}$ and $\trainoutputgt = \left( \futuregt \right)$. From these, the forecaster predicts a future time series $\trainoutputpred \equiv \futurepred$. 
The subscript $b$ in the dataset $\dataset^a_b$ indicates the type of dataset, such as whether it is original or adversarially generated. Likewise, the superscript $a$ indicates if the dataset is from the \textit{train} or \textit{test} distribution.  We compare various forecasters trained on the following datasets, which we call training schemes:
\begin{enumerate}[leftmargin=*]
    \item \Original: The forecaster is \textit{only} trained on $\datasetorigtrain$.  
    \item \DataAdded: We add more training examples from the same distribution as the original training dataset, denoted by $\datasetaddtrain$. This tests whether more examples can improve performance. The task model is then trained on an augmented dataset denoted by $\datasetaddtrain \cup \datasetorigtrain$. 
    \item \Random: We apply zero-mean Gaussian noise with unit variance at each time step to the original training data. The perturbed dataset is denoted by $\datasetrandtrain$ and we re-train the task model on $\datasetrandtrain \cup \datasetorigtrain$.
    \item \Robust \space (Ours): We use our proposed method to train the forecaster. 
\end{enumerate}
For a fair comparison, the \DataAdded, \Random, and \Robust \space schemes add the same number of new training examples to the original training dataset.

\paragraph*{Experiment Procedure}
In both the experiments, the forecaster is initialized with pre-trained parameters on the original train dataset using the mean-squared error loss between the predicted forecasts, $\futurepred$, and the ground-truth forecasts, $\futuregt$. In both experiments, the forecaster and adversary are trained via alternating gradient steps, using the whole training dataset.
As such, first the forecaster makes a prediction from the perturbed history generated from the adversary, and updates its parameters $\paramforcaster$. Then, the adversary predicts the new perturbed history, uses the updated forecaster parameters to calculate the overall cost $\gamecostsymb$, and updates its parameters $\paramadversary$. We repeat this process until convergence, and subsequently check the necessary and sufficient conditions of \ac{lne} ( \cref{prop:first_order_necessary_condition,prop:second_order_necessary_condition}) to check if the converged parameters $\paramforcaster^*$ and $\paramadversary^*$ constitute a \ac{lne}. 





\subsection{ARIMA Forecasting} \label{ssec:arima}
In order to gain intuition about our method's performance on a small, highly-structured dataset, we first examine time series generated by an ARIMA process. For this experiment we have $n=m=p=1$, $x_{0}=1$, $A=C=Q=R=1$, and $B=-1$ and time horizon of $T=50$. The historical time series is of horizon $H=25$ and future time series of horizon $F=25$. Our training dataset is of size $N=4000$ and the test dataset is of size $N=1000$. Thus, the training history time series tensor is $\history \in \mathbb{R}^{4000 \times 1 \times 25}$. The ARIMA time series is: $s_{t+1} = \mu + \alpha s_t + \beta w_{t-1} + w_t$, where $\mu$ is the mean, $w = \mathbb{N}(0, \sigma)$ is noise and $\alpha$, $\beta$ are model parameters. $\mu$, $\alpha$, $\beta$ are initialized randomly. The white noise variance, $\sigma$, is $0.01$ for the original dataset and $0.05$ for the \ac{ood} test dataset. As such, both the original and the \ac{ood} test datasets are generated from quite different distributions. Both hyperparameters $\lambda_f$ and $\lambda_a$ were set to $2.0$ and were chosen experimentally to balance forecaster performance with control performance while also considering nontrivial adversarial perturbations. Our robust game for the ARIMA forecaster converged to a \ac{lne} in approximately 500 iterations.

\subsection{Lane-Change Forecasting} \label{ssec:lane_change}

Now, we demonstrate our method's real world applicability on a challenging lane change dataset \cite{simulator} used to train self-driving policies. This dataset contains 1105 human-human interactive lane change trials from over 19 volunteer drivers in a driving simulator. 
The drivers had to swap lanes with each other within \SI{135}{\meter} of straight road.
The state of each vehicle is $x_t = \left[p_x, p_y, v_x, v_y \right] \in \mathbb{R}^{4}$, where $p_x$ and $p_y$ are the 2-D position of the car in meters and $v_x$ and $v_y$ are the 2-D velocity of the car in $\si{\meter\per\second}$. The control variable is $u_t = \left[a_x, a_y \right] \in \mathbb{R}^{2}$, where $a_x$ and $a_y$ are the 2-D acceleration of the car in $\si{\meter\per\second\squared}$. Each scenario begins with initial conditions drawn randomly. Our training dataset is of size $N=500$ and the test dataset is of size $N=100$.

In this experiment, the forecaster's goal was to predict the ego vehicle's future trajectory in order to complete a successful lane-change, given the history of states of both cars. The historical time series is of horizon $H=20$ and future time series of horizon $F=20$. As such, the forecaster's history time series tensor is $\history \in \mathbb{R}^{500 \times 8 \times 20}$, since it contains the history of time series of both cars. The forecaster's future time series tensor is $\futuregt \in \mathbb{R}^{500 \times 4 \times 20}$. In order to model the measurement noise and uncertainty around the other car's decision making, we restricted the adversary to only be able to perturb the other car's observed historical time series. Therefore, the adversary took the other car's historical state trajectory as input and generated an adversarially perturbed history for that car. The adversary's historical time series tensor is $\history \in \mathbb{R}^{500 \times 4 \times 20}$, since it contains historical time series of only the other car. For the train and test datasets, we used state trajectories with $v_x, v_y \leq \SI{35}{\meter\per\second}$. 
\begin{figure*}[t]
    \begin{center}
    \includegraphics[width=1\textwidth]{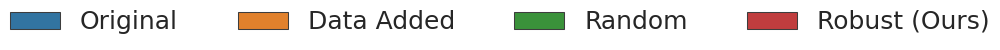}
    \includegraphics[width=1\textwidth]{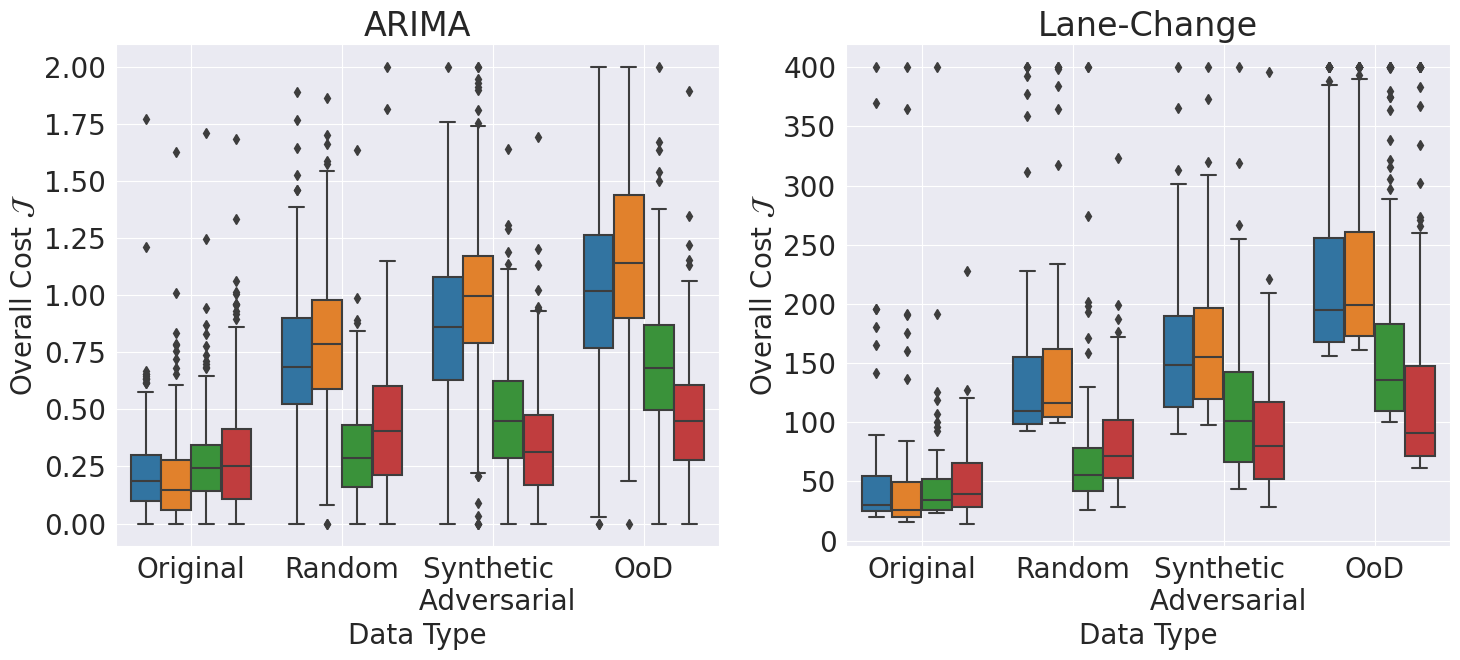}
    \end{center}
    \caption{\small{\textbf{Benefits of game-based training for ARIMA and Lane Change Dataset:} We show the overall cost $\mathcal{J}$ for all training schemes on held-out test environments. The x-axis shows different test conditions, and lower overall cost is better. Our \Robust \space scheme (red) works on par with other forecasters on the \Original \space test data for both experiments. However, it significantly outperforms other forecasters on challenging scenarios in the \textit{out-of-distribution (OoD)} dataset (high-speed lane changes), and \textit{Synthetic Adversarial} test datasets. We beat the \Random \space baseline (green), trained on additional random perturbations, which is not able to model real-world OoD scenarios.}} 
    \label{fig:combined_metrics}
\vspace{-1.2em}
\end{figure*}    

The control policy $\pi$ tracks the predicted future trajectory from the forecaster with a quadratic cost function. Additionally, $A \in \mathbb{R}^{n \times n}, B \in \mathbb{R}^{n \times m}$ matrices in $\dynamics$ follow second-order linear dynamics and $Q \in \mathbb{R}^{n \times n}, R \in \mathbb{R}^{m \times m}$ matrices in the state cost are identity matrices. For the \ac{ood} dataset $\datasetoodtest$, we used real state trajectories with $v_x, v_y > \SI{35}{\meter\per\second}$. The \ac{ood} dataset represents scenarios not seen in the training distribution, but are still possible in the real world and therefore our forecaster should be robust to them. Both hyperparameters $\lambda_f$ and $\lambda_a$ were set to $10.0$ and were chosen experimentally to balance forecaster performance with control performance, while also allowing significant adversarial perturbations. Our robust game for the lane-change forecaster converged to a \ac{lne} in approximately 2000 iterations.

\subsection{Results} \label{ssec:results}
Our experiments show that our robust forecaster training method reduces the overall cost $\mathcal{J}$ compared to benchmarks. 

\textit{Qualitative Results: }
In \cref{fig:adv_scenarios}, we qualitatively demonstrate our method's performance on two \ac{ood} lane change scenarios. Specifically, we show trajectories which are forecasted by models trained using the \Original \space (blue), \Random \space (green), and \Robust \space (red) training schemes given the historical time series. None of the training schemes were exposed to these \ac{ood} scenarios at train time. In these scenarios, two cars are completing a lane-change maneuver. The ground-truth trajectories are shown as dotted lines. All the forecasters were given the same historical time series (time series before the car locations) of both cars to predict the future time series of the ego-vehicle. The control cost $\controlcost$ of each forecasted trajectory is shown in the legend. Both the \Original \space and \Random \space training schemes are not able to correctly predict the ego-vehicle future trajectory and thus lead to higher control costs. Our \Robust \space training scheme is able to correctly predict the ego-vehicle future trajectory and has the lowest control cost.

\textit{Quantitative Results: } \Cref{fig:combined_metrics} shows the overall cost for all schemes on \textit{test} datasets. On the original test dataset, our \Robust \space scheme (red) performs on par with the \Original \space (blue) scheme and slightly worse than the \DataAdded \space (orange) and \Random \space (green) schemes. The \DataAdded \space scheme's performance is expected because it is trained on more original data than our \Robust \space scheme, which allows it to perform slightly better on the original test dataset.

However, the key benefits of our approach are shown on the synthetic adversarial test dataset and held-out \ac{ood} dataset. We run the final trained adversary with final parameters $\paramadversary^*$ on the held-out original test dataset to generate unseen adversarial scenarios which form the synthetic adversarial test dataset. The poor performance of all training schemes on the synthetic adversarial test dataset confirms that the adversary has learned perturbations which are hard for the forecaster, leading to higher control cost. However, our \Robust \space scheme (red) performs significantly better since it was trained to anticipate such unseen perturbations.

In the case of the naturally occurring \ac{ood} test dataset, our \Robust \space training scheme achieves $\mathbf{50.81\%}$ better performance compared to the \Random \space scheme on the ARIMA dataset. Additionally, on the lane-change dataset, our \Robust \space training scheme achieves $\mathbf{30.14\%}$ better performance compared to the \Random \space scheme on the \ac{ood} test dataset. The results show that our \Robust \space training scheme is able to learn robustness to adversarial scenarios and \ac{ood} scenarios. These results are statistically significant using the Wilcoxon signed-rank test \cite{GVK24551600X} with $p<0.05$ for both datasets. The \Random \space scheme is able to generalize better to \ac{ood} data, but is not able to match our \Robust \space scheme's performance, since the random perturbations are relatively benign compared to the targeted scenarios generated by our algorithm. The performance of our \Robust \space training scheme highlights how our game formulation helps the forecaster  generalize better to \ac{ood} scenarios. The \DataAdded \space scheme has overfit to the original dataset, and consequently performs worse than the \Original \space scheme on all other test conditions.



\subsection{Conclusion} \label{sec:conclusion}
In this paper, we considered the challenge of reliable forecasting in robotic decision making. 
We formulated this problem within the framework of a mathematical game played between a learned forecasting model and a hypothetical adversary which may corrupt prior sensor measurements.
Then, we proposed a training scheme which identifies local Nash solutions in this game, and thereby generates more robust forecasts that extend high-quality control performance to unseen \ac{ood} scenarios. 


Despite the existence of theoretical convergence properties for such problems, we observe that convergence is slow in practice, and sensitive to hyperparameters $\lambda_a, \lambda_f$. 
Future work should investigate whether structured parameterizations of the forecaster and adversary might permit reliable convergence to a global Nash Equilibrium. Additionally, we will also validate our method on more complicated forecasters, such as LSTM \cite{Hochreiter1997Long} and Transformers \cite{Vaswani2017Attention}.


\clearpage
\textbf{Acknowledgements:} This material is based upon work supported in part by the Office of Naval Research (ONR) under Grant No. N000142212254. We also gratefully acknowledge the support of the Lockheed Martin AI Center and Viavi Solutions for this research. Any opinions or findings expressed in this material are those of the author(s). They do not necessarily reflect the views of ONR, the Lockheed Martin AI Center, or Viavi.
\balance
\printbibliography

\end{document}